\documentclass[letterpaper]{article} % DO NOT CHANGE THIS
\usepackage{aaai25}  % DO NOT CHANGE THIS
\usepackage{times}  % DO NOT CHANGE THIS
\usepackage{helvet}  % DO NOT CHANGE THIS
\usepackage{courier}  % DO NOT CHANGE THIS
\usepackage[hyphens]{url}  % DO NOT CHANGE THIS
\usepackage{graphicx} % DO NOT CHANGE THIS
\usepackage{natbib}  % DO NOT CHANGE THIS AND DO NOT ADD ANY OPTIONS TO IT
\usepackage{caption} % DO NOT CHANGE THIS AND DO NOT ADD ANY OPTIONS TO IT
\usepackage{algorithm}
\usepackage{algorithmic}
\usepackage{amsfonts}
\usepackage{amsmath}
\usepackage{multirow}
\usepackage{booktabs}
\usepackage{xcolor}
\usepackage{subfigure}
\usepackage{tabularx}
\usepackage{newfloat}
\usepackage{listings}
\DeclareCaptionStyle{ruled}{labelfont=normalfont,labelsep=colon,strut=off} % DO NOT CHANGE THIS
\floatstyle{ruled}
\newfloat{listing}{tb}{lst}{}
\floatname{listing}{Listing}
\title{Detecting and Mitigating Hallucination in Large Vision Language Models via Fine-Grained AI Feedback}
\author{
    Wenyi Xiao\textsuperscript{\rm1}\equalcontrib,
    Ziwei Huang\textsuperscript{\rm1}\equalcontrib,
    Leilei Gan\textsuperscript{\rm1}\thanks{Corresponding author.},
    Wanggui He\textsuperscript{\rm2}\\
    Haoyuan Li\textsuperscript{\rm2},
    Zhelun Yu\textsuperscript{\rm2},
    Fangxun Shu\textsuperscript{\rm2},
    Hao Jiang\textsuperscript{\rm2},
    Linchao Zhu\textsuperscript{\rm1}
}
\affiliations{
    \textsuperscript{\rm 1}Zhejiang University\\
    \textsuperscript{\rm 2}Alibaba Group\\
    \{wenyixiao, leileigan\}@zju.edu.cn,
}

\usepackage{bibentry}
\begin{document}

\maketitle

\begin{abstract}
The rapidly developing Large Vision Language Models (LVLMs) still face the \textit{hallucination phenomena} where the generated responses do not align with the given contexts, significantly restricting the usages of LVLMs. Most previous work detects and mitigates hallucination at the coarse-grained level or requires expensive annotation (e.g., labeling by human experts or proprietary models). To address these issues, we propose detecting and mitigating hallucinations in LVLMs via fine-grained AI feedback. The basic idea is that we generate a small-size sentence-level hallucination annotation dataset by proprietary models, whereby we train a detection model which can perform sentence-level hallucination detection. Then, we propose a detect-then-rewrite pipeline to automatically construct preference dataset for hallucination mitigation training. Furthermore, we propose differentiating the severity of hallucinations, and introducing a Hallucination Severity-Aware Direct Preference Optimization (HSA-DPO) which prioritizes the mitigation of critical hallucination in LVLMs by incorporating the severity of hallucinations into preference learning. Extensive experiments on hallucination detection and mitigation benchmarks demonstrate that our method sets a new state-of-the-art in hallucination detection on MHaluBench, surpassing GPT-4V and Gemini, and reduces the hallucination rate by 36.1\% on AMBER and 76.3\% on Object HalBench compared to the base model.
\end{abstract}

\begin{links}
    \link{Code}{https://github.com/Mr-Loevan/HSA-DPO}
\end{links}

\section{Introduction}
\begin{figure}[t]
    \centering
    \includegraphics[width=0.9\linewidth]{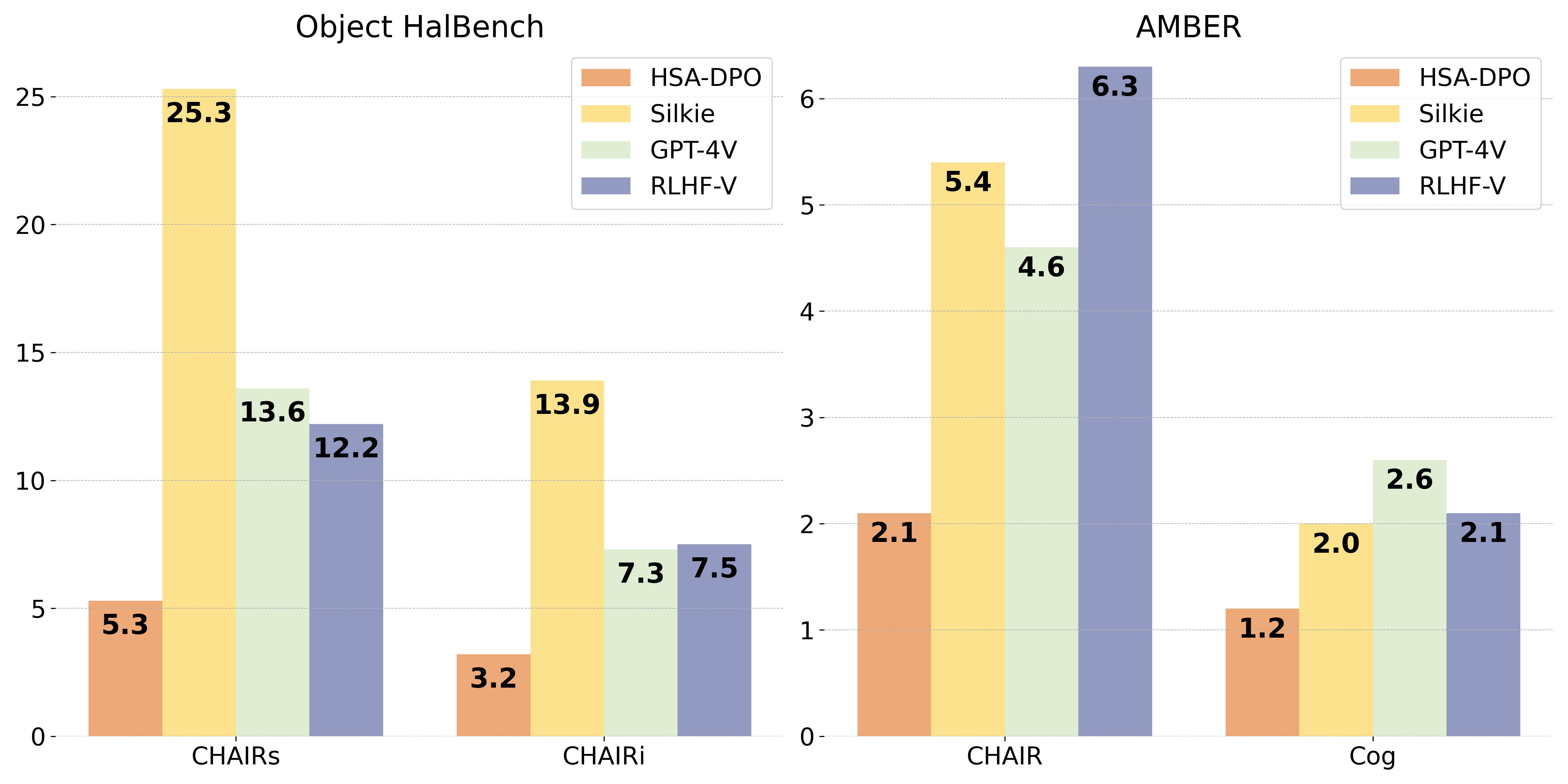}
    \caption{Comparison of HSA-DPO (red) with state-of-the-art models in mitigating hallucinations (Silkie, GPT-4V, RLHF-V) on Object HalBench and AMBER benchmarks. Notably, HSA-DPO outperforms state-of-the-art models in all metrics.}
    \label{fig:enter-label}
\end{figure}
Large Language Models (LLMs)~\cite{chatgpt, llama, GPT-4, mistral} have marked a significant milestone in the development of natural language processing and have been further extended to encompass multi-modality data, such as language and vision, leading to the emergence of Large Vision Language Models (LVLMs)~\cite{gpt-4V, llava, gemini, qwen-vl}. Despite the remarkable performance of LVLMs across a broad spectrum of vision-language tasks~\cite{shikra,llava-1.5,minigpt-4,chen2023internvl,instructblip, CMSCGC, SSGCC}, LVLMs still grapple with the phenomena of \textit{hallucination}, wherein the completions do not align with the given contexts. In other words, the generated responses contain incorrect objects, attributes and relations concerning the vision and language inputs, thereby significantly restricting the utility of LVLMs~\cite{liu2024survey,yin2023survey}. 

Research on addressing hallucinations in LVLMs can be primarily categorized into hallucination detection and mitigation. Hallucination detection aims to identify the presence of hallucinations in the LVLM outputs for preventing potential malicious usages~\cite{pope, amber, objecthalbench, llava-rlhf, lrv, unified}. 
%For example, UNIHD~\cite{unified} leverages external visual tools to detect whether hallucinations exist in each claim of hallucinatory response. 
Hallucination mitigation aims to enable LVLMs to generate more faithful responses and can be mainly divided into training-free and training-based approaches~\cite{woodpecker, opera, llava-rlhf, ha-dpo,rlhf-v,fdpo,zhou2024povid}. Training-free approaches address potential hallucinations by post-processing the outputs of LVLMs~\cite{woodpecker,opera,zhou2023lure,han2024skip}. While not requiring additional training costs, training-free approaches tend to reduce the inference speed. On the other hand, training-based approaches seek to reduce hallucinations in LVLMs through further instruction fine-tuning~\cite{lrv,yue2024less} or preference learning~\cite{llava-rlhf,rlhf-v,silkie,ha-dpo,fdpo} on specifically constructed datasets. Some recent studies~\cite{silkie,ha-dpo,fdpo} have exploited feedback from powerful closed-source LVLMs for improving the fidelity of LVLMs responses.

However, despite the aforementioned efforts, several challenges persist in detecting and mitigating hallucinations in LVLMs. First, the preference data is generally at response-level~\cite{llava-rlhf,silkie,ha-dpo,fdpo}, which is sub-optimal for thoroughly detecting and mitigating hallucinations. Second, constructing preference data for training-based mitigation approaches requires expensive annotations either by human experts~\cite{llava-rlhf,rlhf-v} or proprietary commercial models~\cite{silkie,ha-dpo,fdpo,unified,muffin}, especially if fine-grained annotation is involved. Lastly, existing studies often treat all hallucinations equally, leading to scenarios where less significant hallucinations are addressed, while more critical ones are neglected. For example, in certain scenarios, compared to incorrect color descriptions of objects, addressing the hallucinatory description of non-existent objects should be prioritized.

\begin{figure*}[t]
  \centering
  \includegraphics[width=0.92\textwidth]{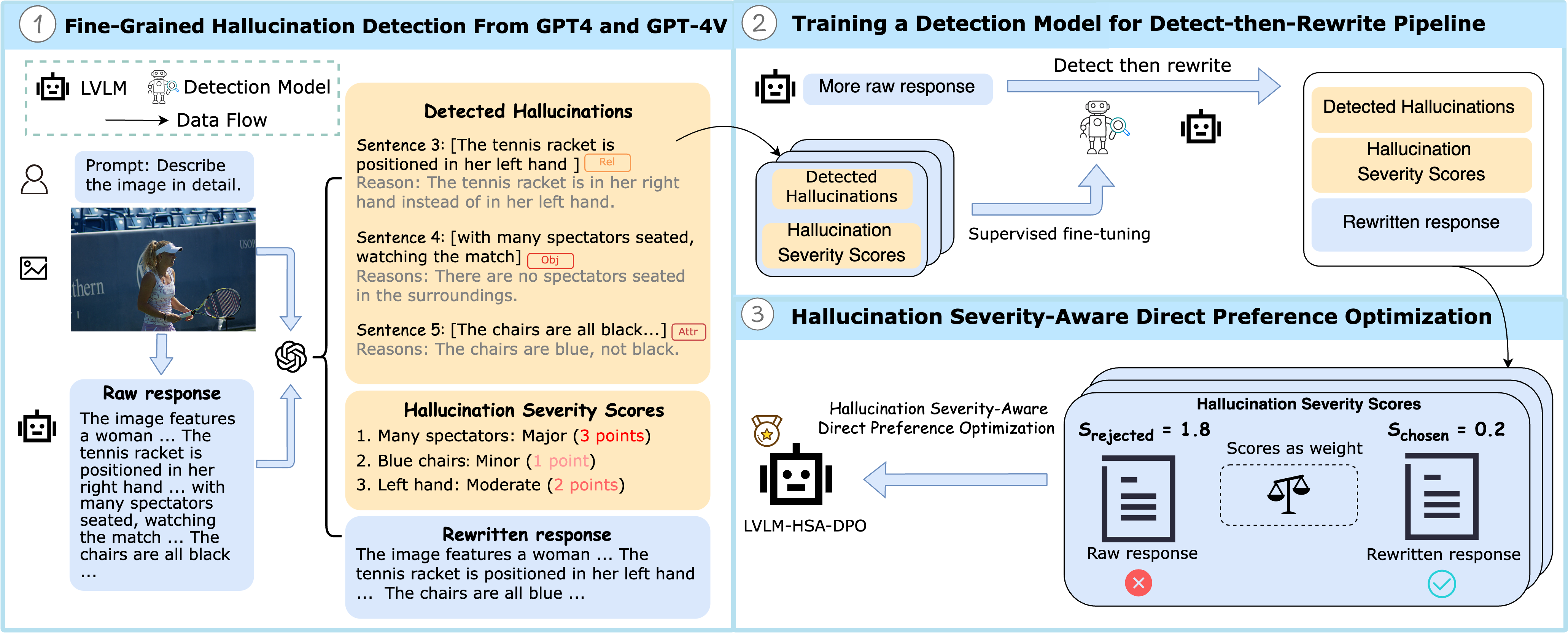}
  \caption{Our work consists of three components: \S~\ref{sec:fine_grained_ai_feedback} fine-grained hallucination detection from GPT-4/GPT-4V; \S~\ref{sec:hallucination_detection} hallucination detection model for detect-then-rewrite preference dataset construction pipeline; and \S~\ref{sec:hallucination_mitigation} hallucination severity-aware direct preference optimization.}
  \label{fig:bigframework}
\end{figure*}

To address these issues, in this work, we propose detecting and mitigating hallucinations in LVLMs via fine-grained AI feedback. As shown in Figure~\ref{fig:bigframework}, our framework consists of three key components: (1) {\bf Fine-Grained AI Feedback.} The initial step involves generating a small-scale, sentence-level hallucination annotation dataset by proprietary models. Beyond merely detecting hallucinations, we meticulously craft prompts to collect detailed feedback on the type, severity and rationale of each hallucination. Compared to coarse-grained feedback, this sentence-level granularity ensures more precise and thorough hallucination detection. (2) {\bf Fine-Grained AI Feedback for Hallucination Detection.} The next step proposes training a hallucination detection model using this fine-grained AI feedback, enabling it to perform sentence-level hallucination detection across primary types (e.g., object, attribute, and relationship). This step also introduces an automatic pipeline for constructing preference dataset where given a hallucinatory response, the detection model first identifies hallucinations within each sentence of the response. Based on the detected hallucinations, a rewriting model then revises the hallucinatory response into non-hallucinatory one, forming the $<$chosen\_answer, rejected\_answer$>$ pair. This pipeline enables us to more cost-effectively annotate a large-scale preference dataset for training mitigation models. The underlying insight behind this approach aligns with the concept of scalable oversight which aims to train machines to assist humans in accurately evaluating model output~\cite{constitutionalAI, rlaif,ganguli2023capacity,mcaleese2024llm}. (3) {\bf Hallucination Severity-Aware DPO.} Lastly, we propose differentiating the severity of hallucinations, and introduce a Hallucination Severity-Aware Direct Preference Optimization (HSA-DPO). HSA-DPO incorporates hallucination severity into preference learning for prioritizing the mitigation of critical hallucinations. 

We conduct extensive experiments on a range of hallucination detection and mitigation benchmarks and the experimental results have demonstrated the effect of the proposed method. For hallucination detection, our detection model achieves new state-of-the-art results on MHaluBench, surpassing GPT-4V and Gemini. As shown in Figure~\ref{fig:enter-label}, for hallucination mitigation, HSA-DPO improves the base LVLM by reducing the Hallucination Rate on AMBER by 36.1\% and $\text{CHAIR}_{S} $ on Object HalBench by 76.3\%. These results demonstrate the effectiveness of fine-grained AI feedback and the proposed HSA-DPO.

\section{Related Work}

In this section, we give a brief introduction of related studies on LVLMs hallucination detection and mitigation.
% \vspace{-5pt}
\subsection{Detecting Hallucination in LVLMs}

Current approaches for hallucination detection mainly focus on utilizing the abilities of off-the-shelf tools, such as closed-source LLMs, LVLMs or visual tools.  GAVIE~\cite{lrv} employs GPT-4 to facilitate the evaluation of object hallucinations. \citet{ha-dpo} introduce the sentence-level hallucination metric SHR, which harnesses GPT-4 to determine the presence of hallucinations in LVLM outputs. UNIHD leverages GPT-4V~\cite{gpt-4V} or Gemini~\cite{gemini} to extract verifiable claims from the generations of LVLMs, and then uses visual tools for hallucination detection. Compared to previous studies, our detection model is trained on fine-grained feedback from proprietary LVLMs, which covers main hallucination types (i.e., object, attribute, and relationship). It can also evaluate the severity of hallucinations and provide detailed reasons.

\subsection{Mitigating Hallucination in LVLMs}
Hallucination mitigation can be mainly divided into training-free and training-based approaches~\cite{woodpecker, opera, llava-rlhf, ha-dpo,rlhf-v,fdpo,fgaif}. Training-free approaches address potential hallucinations by post-processing the outputs of LVLMs~\cite{woodpecker,opera}, thereby tending to reduce the inference speed of LVLMs. 
%For example, Woodpecker~\cite{woodpecker} introduces a detect-then-correct hallucination mitigation approach with the aid of ChatGPT and a VQA model~\cite{blip-2}. 
%However, training-free approaches tend to reduce the inference speed. 
Instead, the latter reduce hallucinations in LVLMs via further training, such as instruction fine-tuning~\cite{lrv} or preference learning~\cite{llava-rlhf,rlhf-v,silkie,ha-dpo,fdpo,fgaif}. 
%For example, LRV\cite{lrv} performs length-controlled fine-tuning on visual instructions to alleviate hallucinations.
Our work belongs to preference learning which biases LVLMs to favor the non-hallucinatory responses~\cite{ha-dpo,fdpo,llava-rlhf,rlhf-v}. LLaVA-RLHF~\cite{llava-rlhf} is the first to train an LVLM to align with human preference. 
RLHF-V~\cite{rlhf-v} manually collects segment-level human preference and conduct dense DPO over the human feedback to reduce hallucinations. 
POVID~\cite{zhou2024povid} constructs preference dataset by inserting textual hallucinations and distorting input images.
%FDPO~\cite{fdpo} proposes a fine-grained DPO, which does not rely on pair-wise preference dataset.
Silkie~\cite{silkie} exploits feedback from various LVLMs for constructing preference dataset, but the feedback are coarse-grained. In this study, we propose a pipeline for automatically constructing preference dataset and introduce the hallucination severity-aware preference learning for prioritizing the mitigation of critical hallucinations.

% \vspace{-5pt}
\section{Methodology}
\label{sec:method}
%The core idea of our method is to replace human annotation with fine-grained AI feedback for detecting and mitigating LVLM hallucination. 
In this section, we begin by introducing how to gather fine-grained AI feedback in \S\ref{sec:fine_grained_ai_feedback}. Following this, we detail how this fine-grained AI feedback is used for detecting and mitigating LVLM hallucinations in \S\ref{sec:hallucination_detection} and \S\ref{sec:hallucination_mitigation}, respectively.

\subsection{Fine-Grained AI Feedback Generation}\label{sec:fine_grained_ai_feedback}
Before introducing the method for gathering fine-grained AI feedback, we first detail the process of generating hallucinatory responses, upon which the collection of AI feedback is performed.
\paragraph{Hallucinatory Response Generation} We investigate hallucination in the tasks of Detailed Description Generation (DDG) and Visual Complex Reasoning (VCR) following~\citet{lrv}. These tasks require the LVLM to generate longer detailed description or complex reasoning response given the visual-language content, making the LVLM more susceptible to produce hallucinations. Note that our method is not limited to the two tasks but can be extended to other visual-language tasks. 

We choose the Visual Genome (VG)~\cite{visualgenome} and Silkie~\cite{silkie} dataset for constructing the DDG and VCR prompts, respectively. The images in VG are content-rich and associated with bounding boxes which specify various objects, attributes of each object, and spatial relationships within image content. These detailed annotations can help obtain more accurate AI feedback. 
% For DDG, we randomly select 5,000 images from the VG dataset and acquire the detailed object bounding boxes annotation for each image. For VCR, following~\citet{lrv}, we randomly select 1,000 <image, question> pairs from the VG dataset. The combination of the two tasks results in a total of 6,000 instances upon which we generate, denoted as $\mathcal{D}_\text{prompt}$.
% With these elaborate annotations, it is possible for LVLMs to give more accurate feedback on hallucination detection. 
% Specifically, for DDG, we randomly select 5,000 images from the VG dataset and acquire the detailed object bounding boxes annotation for each image. For VCR, following~\citet{lrv}, we randomly select 1,000 <image, question> pairs from the VG dataset. The combination of the two tasks results in a total of 6,000 instances to, denoted as $\mathcal{D}_\text{prompt}$. 
Specifically, given a target LVLM $M$, for DDG, a randomly selected image from VG and an instruction from the instruction set in RLHF-V\cite{rlhf-v} are used as the prompt for $M$ to generate a potentially hallucinatory response. Randomly choosing instruction from the instruction set injects randomness and harnesses the model to discern intricate image details. For VCR, we randomly select a $<$image, question$>$ pair from the Silkie dataset as the prompt to generate a potentially hallucinatory response. 

% Consequently, we obtained a total of 6,000~\textbf{Hallucinatory Responses} produced by the target LVLM $M$. Next, we can collect fine-grained feedback from GPT-4 or GPT-4V on these hallucinatory responses.

%This method yielded 5k hallucinatory responses for detailed description generation.

We denote the set of generated hallucinatory responses as $\mathcal{D}_\text{hal} = \{(x_i, \hat{y}_i)\}_{i=1}^N$ where $\hat{y}_i$ is the hallucinatory response, $x_i$ is the corresponding prompt and $N$ is the size of $\mathcal{D}_\text{hal}$.

\paragraph{Fine-Grained Hallucination Annotation via GPT-4/GPT-4V.}

Given the collected hallucinatory dataset $\mathcal{D}_\text{Hal}$, we can now gather fine-grained AI feedback upon it using GPT-4 and GPT-4V. The motivation behind this is that manually annotating large-scale datasets at fine-grained level is time-consuming, costly, and challenging. 

% Specifically, for DDG, we input the hallucinatory response and the associated verbalized object bounding boxes into GPT-4 to generate feedback. The reason for it is that the object bounding boxes provide a more accurate content compared to the original image. As for visual complex reasoning, the inputs consisted of the original images along with the LVLM-generated hallucinatory response. As shown in Figure~\ref{fig:framework_detail}, feedback from GPT-4 and GPT-4V consists of the following aspects.
% \textbf{(1) Sentence-level hallucinations with explanations.} The first step involves identifying hallucinations in response $\hat{y}$ at a rigorous sentence-level. 

Specifically, for each VCR hallucinatory sample $(x_i, \hat{y}_i)$ in $\mathcal{D}_\text{Hal}$, we input it into GPT-4V to generate fine-grained AI feedback at a rigorous sentence-level. For DDG, we provide $(x_i, \hat{y}_i)$ along with the associated verbal object bounding boxes in the VG dataset to GPT-4 as these additional annotations enable more accurate AI feedback compared to relying solely on the images. The used instruction prompt to generate the fine-grained AI feedback is shown in Figure 1 and 2 of the supplementary material. The obtained feedback is a six-tuple $(x_i, \hat{y}^{j}_i, h^j_{i,\text{type}}, h^j_{i,\text{R}}, \text{HS}^j_i, \text{HS}^j_{i, \text{R}})$ where $\hat{y}^j_i$ is the $j$-th sentence of $\hat{y}_i$, $h^j_{i,\text{type}}$ is the hallucination type, $h^j_{i,\text{R}}$ is the reason that explains why $\hat{y}^j_i$ is considered a hallucination, $\text{HS}^j_i$ is the hallucination severity score used to differentiate the effect of different hallucinations and $\text{HS}^j_{i, \text{R}}$ is the reason for hallucination severity score $\text{HS}^j_i$. The provided reasons $h^j_{i,\text{R}}$ and $\text{HS}^j_{i, \text{R}}$ improve the explainability of the hallucination detection process. Compared to coarse-grained feedback, this sentence-level granularity ensures a thorough hallucination detection. 

For $h^j_{i,\text{type}}$, we consider the following types of hallucinations: (i) $<$object$>$ for object hallucinations, such as perceiving physical entities that are not actually present; (ii) $<$relationship$>$ for relationship hallucinations, such as giving inaccurate description of the relationship between objects; (iii) $<$attribute$>$ for attribute hallucinations, such as inaccurate perceptions of the characteristics of objects. For the hallucination severity score $\text{HS}^j_i$, we define the following Likert-style ratings: (i) Minor (1 point): the hallucination concerns a minor detail and does not significantly affect the overall portrayal of the scene; (ii) Moderate (2 points): the hallucination involves a noticeable detail that is incorrect within the context of the scene, yet the overall comprehension of the scene is maintained; (iii) Major (3 points): the hallucination introduces a significant error or an entirely fabricated element that fundamentally alters the viewer's understanding of the scene. These scores facilitate the assessment of hallucination severity and are further incorporated into the preference learning(See \S\ref{sec:hallucination_mitigation}) to prioritize the mitigation of critical hallucinations. We denote the set of fine-grained AI feedback as $\mathcal{D}_\text{faif}$.

\subsection{Hallucination Detection via Fine-Grained AI Feedback}
\label{sec:hallucination_detection}
With the collected fine-grained AI feedback dataset $\mathcal{D}_\text{faif}$, we can then train a hallucination detection model using open-source LVLMs. Training an open-source model for hallucination detection offers the following merits. First, it enables perform fine-grained LVLMs hallucination detection with severity scores but not relies on proprietary models at lower cost. Second, the detected fine-grained hallucinations can be further used to construct a preference dataset, as discussed in $\S\ref{sec:hallucination_mitigation}$.

% dataset denoted as $\mathcal{D}_{sft}$ to Supervised Fine-Tuning (SFT) dection model $M_d$
% In practice, We segmented the 6k Detected hallucinations into individual sentences, differentiating hallucinated sentences and non-hallucinated sentences. For hallucinated sentences, we paired them with explanations for the hallucination and scores of the hallucination severity.
% $
% \mathcal{D}_{\text{sft}} = \{(\text{Question}, \text{Response})\}, \text{ where Question} = (\text{Prompt}, \text{Sentence}), \text{ and Response} = (\text{Reasons}, \text{Scores})
% $

Formally, given the sentence-level training dataset $\mathcal{D}_\text{faif}=\{(x^i, \hat{y}^i, h^i_\text{type}, h^i_\text{R}, \text{HS}^i, \text{HS}^i_\text{R})\}_{i=1}^M$, we train fine-grained hallucination detection model $M_\text{det}(;\theta)$, parameterized by $\theta$, by minimizing the negative log likelihood loss: 
\begin{equation}
\label{eq:sft}
    \mathcal{L}_\text{DET}(\theta)=-\sum_{i=1}^{M}\sum_{t=1}^{T}\log M_{det}(g^i_t|x^i,\hat{y}^i,g^i_{1:t})
\end{equation}
where $g^i$ is concatenation of $h^i_\text{type}, h^i_\text{R}, \text{HS}^i, \text{HS}^i_\text{R}$. Note that for non-hallucinated sentences, $h^i_\text{type}, h^i_\text{R}, \text{HS}^i, \text{HS}^i_\text{R}$ are set to $<$No hallucination, None, 0, None$>$, respectively. Our preliminary experiments indicate that the ratio of hallucinated to non-hallucinated data significantly impacts the performance of the detection model. We tested multiple ratios and found that a final ratio of 1:1.2 provides optimal results.

\subsection{Hallucination Mitigation via Fined-Grained AI Feedback}
\label{sec:hallucination_mitigation}
% In this section, we introduce the hallucination severity-aware direct preference optimization for mitigating the hallucinations in the target LVLM, where the preference dataset is constructed by an automated detect-then-rewrite pipeline.
%Our ultimate goal is to mitigate the hallucination problem in the target LVLM. To achieve this
With the built hallucination detection model, we next introduce an automated method for constructing the preference dataset.
\paragraph{Detect-then-Rewrite Pipeline for Automatic Preference Dataset Construction.} 
% As mentioned in the introduction, constructing preference dataset to mitigate hallucination typically requires expensive annotations, either by human experts or proprietary commercial models, and the preference data are generally response-level~\cite{llava-rlhf,silkie,ha-dpo,fdpo}, which is sub-optimal for thoroughly detecting and mitigating hallucinations. To address this, we propose a detect-then-write pipeline for automatically constructing the preference dataset. 
To reduce the expensive annotation costs either caused by human experts or proprietary models~\cite{llava-rlhf,silkie,fdpo}, we propose a detect-then-write pipeline for automatic preference dataset construction, which allows for cost-effectively fine-grained feedback annotation at scale. 
%The efficiency and cost of annotation using our pipeline compared to previous methods are provided in Table~\ref{tab:complexity}.

Specifically, the detect-then-rewrite pipeline consists of the hallucination detection model, $M_\text{det}$, and one rewriting model, $M_\text{wri}$. Given a prompt and its hallucinatory response $(x, \hat{y})$, the detection model first identifies a set of hallucination $\mathcal{H}=\{( h^j_\text{type}, h^j_\text{R}, \text{HS}^j, \text{HS}^j_\text{R})\}_{j=1}^{|\hat{y}|}$. Then, using $\hat{y}$ and $\mathcal{H}$ as the input prompt, $M_\text{det}$ rewrites $\hat{y}$ into a non-hallucinatory response $y$. The specific rewriting prompt is provided in section A.1 of the supplementary material. In practice, we choose an open-source LVLM LLaVA as the rewriting model $M_w$, as it not only has demonstrated the impressive instruction-following and rewriting capability in our pilot experiments, but also offers a way to further reduce the annotation cost. We denote this preference dataset as $\mathcal{D}_\text{pref}=\{(x_i, \hat{y}_i, y_i, \mathcal{H}_i)\}_{i=1}^{N}$ where $N$ is the dataset size.

\paragraph{Connection to Scalable Oversight.} Compared with previous studies, this preference dataset construction pipeline enables us to more budget-friendly annotate a large-scale preference dataset for training mitigation models. The underlying insight behind this approach is closely correlated with the concept of \emph{scalable oversight}, which aims to train machines to assist humans in supervising models by critiquing the model's output~\cite{constitutionalAI, rlaif,ganguli2023capacity,mcaleese2024llm} or decomposing complex problems into simpler sub-problems\cite{leikeScaleable,DBLP:conf/iclr/LightmanKBEBLLS24}. In this work, we break down the complicated fine-grained labeling process into two steps: first, detecting (critiquing) hallucinations in the generated completion, and then rewriting them into non-hallucinated ones. Moreover, we leverage the capabilities of current open-source LVLMs by employing them as hallucination detection and rewriting experts, thereby reducing the cost of providing a large-scale supervisions. Notably, our pipeline does not require any ground truth datasets, making it not only cost-effective but also practical for curtain scenarios where labeled datasets may be unavailable.

\paragraph{Hallucination Severity-Aware Direct Preference Optimization.}  
With the automatically constructed preference dataset, we can now perform preference learning for hallucination mitigation. In particular, we choose the offline preference optimization method Direct Preference Optimization (DPO)~\cite{rafailov2024direct,xiao2024comprehensivesurveydirectpreference} as it is more stable and efficient compared to online RLHF methods\cite{ouyang2022training,ppo}. The learning objective of DPO is directly formulated over the the policy model $\pi_\theta (y|x)$ and a reference model $\pi_\text{ref} (y|x)$:

\begin{equation}
\footnotesize
\mathcal{L}_\text{DPO} = -\mathop{\mathbb{E}}_{(x, y_w, y_l) \in \mathcal{D}}\bigl[\log \sigma(\beta\log \frac{\pi_\theta (y_w|x)}{\pi_{\text{ref}} (y_w|x)} - \beta\log \frac{\pi_\theta (y_l|x)}{\pi_{\text{ref}} (y_l|x)})\bigr] \\
\end{equation}
where $y_l$ and $y_w$ are the rejected and chosen answer. $\sigma$ is the logistic function. $\beta$ is the hyper-parameter controlling the deviation from $\pi_\text{ref} (y|x)$. Action score $\log \pi(y|x)$ is the response generation likelihood.

As can be observed, the standard DPO loss $\mathcal{L}_\text{DPO}$ treats all pairwise preference responses equally, making more severe hallucinations (e.g., description of non-existent objects) not being greater considered compared with other hallucinations (e.g., incorrect color descriptions of objects). To tackle this limitation, we present Hallucination Severity-Aware Direct Preference Optimization (HSA-DPO), which incorporates hallucination severity into the preference optimization for mitigating critical hallucinations with higher priority. Specifically, we begin with aggregating the sentence-wise hallucination severity scores in $\mathcal{H}_i$ as the response-level severity score:
% \vspace{-2pt}
\begin{equation}
    S^i_\text{AVG} = \frac{1}{T} \sum_{j=1}^{T} \text{HS}^j_i 
\end{equation}
where $T$ is the number of sentences in the response which helps prevent potential reward hacking introduced by the length bias. This strategy bears similarity to our fine-grained hallucination detection, which can alleviate the difficulty of directly assessing the hallucination severity of the whole response.
Subsequently, we adaptively assign this severity score to the implicit reward model of DPO to ensure responses with more severe hallucinations receive stronger penalties for correction:
\begin{equation}
\footnotesize
\mathcal{L}_\text{MIT} = -\sum_{i=1}^{|\mathcal{D}_\text{pref}|} \log \sigma(\beta\bigl[\log \frac{\pi_\theta (y_i|x_i)}{\pi_{\text{ref}} (y_i|x_i)} - S^i_\text{AVG} \log  \frac{\pi_\theta (\hat{y}_i|x_i)}{\pi_{\text{ref}} (\hat{y}_i|x_i)}\bigr])
\end{equation}

\section{Experiments}
In this section, we evaluate the efficacy of our method for detecting and mitigating hallucinations in LVLMs. 
% In this section, we empirically investigate the effectiveness of H-DR and SA-DPO in detecting and mitigating LVLM hallucinations. We adopt benchmarks that directly evaluate the long-form responses, which are more closely related to the practical usage scenarios of LVLMs. Additionally, we argue that binary classification evaluation(i.e., answering yes/no) also indicate the ability of preventing hallucination. For trustworthiness, we perform evaluation on five benchmarks.
% \vspace{-5pt}
\subsection{Datasets and Metrics}
We introduce the datasets and metrics for evaluating hallucination detection and mitigation.
For hallucination detection, we use the following benchmarks: (1) \textbf{MHaluBench}\cite{unified} is a newly established benchmark for detecting hallucination in both image-to-text and text-to-image settings as binary classification task. We adopt the image-to-text part to evaluate our detection model. (2) To evaluate various types of hallucination and their severity, we manually labeled a dataset named \textbf{MFHaluBench}, which includes object, attribute, and relationship hallucinations along with a human-annotated severity score for each segment. We evaluate MFHaluBench using binary classification to identify hallucinated segments and multi-class classification to distinguish between different types of hallucinations.

\setlength{\tabcolsep}{0.8mm}
\begin{table}[t]
\renewcommand\arraystretch{1.22}
  \centering

  \small
  % \resizebox{0.50\textwidth}{!}
 {
 \begin{tabular}{ccccccc}
\toprule
\multirow{2}{*}{\textbf{Methods}} & \multirow{2}{*}{\textbf{Levels}} 
& \multicolumn{4}{c}{\textbf{Average}}\\
\cline{3-6}
& & \textbf{Acc.} & \textbf{P} & \textbf{R} & \textbf{Mac.F1} \\
\midrule
\multirow{2}{*}{Gemini with Self-Check} & Claim  & 74.74 & 75.80 & 75.68 & 74.74 \\
& Segment  & 75.11 & 73.89 & 77.44 & 73.85 \\
\cline{2-6}
\multirow{2}{*}{\textbf{Gemini with UNIHD}}   & Claim & 77.41 & 77.76 & 77.99 & 77.39 \\
& Segment & 78.68 & 75.97 & 78.64 & 76.74 \\
\cline{1-6}     
\multirow{2}{*}{GPT-4V with Self-Check} & Claim  & 79.25 & 79.02 & 79.16 & 79.08 \\
 & Segment  & 80.80 & 77.80 & 78.30 & 78.04 \\
\cline{2-6}
\multirow{2}{*}{\textbf{GPT-4V with UNIHD}}   & Claim & 81.91 & 81.81 & 81.52 & 81.63 \\
& Segment & 84.60 & 82.77 & 80.89 & 81.71 \\
\cline{1-6}
\multirow{2}{*}{\textbf{Our Detection Model}} & Claim  & \textbf{85.60} & \textbf{85.46} & \textbf{85.79} & \textbf{85.52}\\
& Segment & \textbf{86.94} & \textbf{87.73} & \textbf{82.88} & \textbf{85.23}\\
\bottomrule
\end{tabular}
  }
    \caption{Experimental results of MhaluBench on Image-to-Text hallucination detection. The results for Gemini and GPT-4V are sourced from the UNIHD\cite{unified}.}
    \label{tab:main results in detection}
  \label{tab:allresults}%
\end{table}%

\setlength{\tabcolsep}{1.5mm}
\begin{table}[t]
    \centering
     
    % \resizebox{0.46\textwidth}{!}
    {
        \begin{tabular}{ccccc|c}
        \toprule
        \multirow{2}{*}{\textbf{Model}} & \multicolumn{4}{c}{\textbf{Binary}} & \multicolumn{1}{c}{\textbf{Multi}} \\
        \cline{2-6} 
         & \textbf{P} & \textbf{R} & \textbf{ACC} & \textbf{F1} & \textbf{ACC}  \\
        \midrule
         GPT-4V 2shot & 59.7 & 98.7 & 63.3 & 74.4 & 40.6 \\
         LLaVA-1.6-34B 2shot & 55.5 & 100 & 56.7 & 71.4 & 36.7 \\
        Our detection model & 87.8 & 88.8 & 87.3 & 88.2 & 74.3 \\
        \bottomrule
        \end{tabular}
    }
    \caption{Experimental results of MFHaluBench. Details of Multi refer to section B.3 of the supplementary material.}
     \label{tab:finegrainedresults}
\end{table}

For hallucination mitigation, we use the following benchmarks: (1) \textbf{Object HalBench}~\cite{objecthalbench} is a widely adopted benchmark for evaluating object hallucination in DDG. Following~\citet{rlhf-v}, we use $\text{CHAIR}_{S} $(i.e., the percentage of responses that contain hallucinations) and $\text{CHAIR}_{I}$ (i.e., the percentage of hallucinated object mentions among all object mentions) as the evaluation metrics.
(2) \textbf{AMBER}~\cite{amber} consists of generative and discriminative parts, focusing on common objects and pitfall objects which easily cause hallucinations.  We use the generative part of AMBER and report the following metrics: CHAIR, Cover, Hal and Cog. More details about these metrics can be found in section C.1 of the supplementary material. 
(3) \textbf{MMHal-Bench}~\cite{llava-rlhf} is a benchmark for evaluating object hallucination by using GPT-4 to compare the model output with the annotated response. We report overall score rated by GPT-4 and the hallucination rate.
(4) \textbf{POPE}~\cite{pope} is an object hallucination evaluation benchmark by testing LVLMs in a form of question answering. We choose the Adversarial part of POPE and report its F1 scores.

In addition to the above hallucination detection and mitigation benchmarks, we also adopt the widely used \textbf{LLava Bench in the wild}\cite{llava} to evaluate the multi-modal capabilities after mitigating training. We also introduce \textbf{Hallucination Severity Score}: a metric to evaluate the hallucination severity of model response. Severity scores ranges from 0 to 3. For complete definitions of these scores, refer to section A.2 of the supplementary material.
% \vspace{-5pt}

\setlength{\tabcolsep}{1.6mm}
\begin{table*}
  \small
  % \resizebox{1.0\textwidth}{!}
  {
    \begin{tabular}{lc cc cccc cc ccc}
    \toprule
      \multirow{2}{*}{\textbf{Model}}   & \multicolumn{2}{c}{\textbf{Object HalBench} } & \multicolumn{4}{c}{\textbf{AMBER}} & \multicolumn{2}{c}{\textbf{MMHal-Bench}} & \multicolumn{1}{c}{\textbf{LLaVA Bench}}  & \multicolumn{1}{c}{\textbf{POPE Adv.}}  \\

      \cmidrule(lr){2-3} \cmidrule(lr){4-7} \cmidrule(lr){8-9} \cmidrule(lr){10-10} \cmidrule(lr){11-11}
      
      & $\text{CHAIR}_{S} $$\downarrow$ & $\text{CHAIR}_{I} $ $\downarrow$  & CHAIR $\downarrow$ & Cover. $\uparrow$ & Hal.$\downarrow$ &  Cog.$\downarrow$ & Overall $\uparrow$ & Resp. $\downarrow$ & Overall $\uparrow$ & F1 $\uparrow$  \\

    \midrule
    %   LLaVA~\cite{liu2023visual}        & 63.0 & 29.5 &  46.6   &  21.2  & 19.9 &  80.8 & 31.9 & 70.8  & 85.4  & 74.3 & 96.3  & - \\
    % % Muffin~\cite{yu2023reformulating}       & 50.5 & 24.5 & 33.6 & 16.4 & 26.0 & 74.7 & 33.4 & 68.8 & 89.3 & \textbf{79.7} & \underline{97.7} & - \\
      LRV   & 32.3    & 22.3      & -  & -  & - &- & -  & - & - & - \\
      POVID   & 48.1   & 24.4      & 7.3  & 49.5  & 31.1 & 3.7 & 2.08  & 0.56 & - & 81.6 \\
    % Muffin-QA~\cite{yu2023reformulating}       & 27.4 & 14.9 & 33.6    & 17.1    & 17.8  & 65.1  & \underline{39.6} & 58.3 & \underline{94.9}  & 76.1 & 90.9   & \textbf{80.0} \\
      InstructBLIP   & 25.9 & 14.3  & 8.8  & 52.2  & 38.2 & 4.4  & 2.14 & 0.58 & - & 78.4   \\
      Qwen-VL-Chat & 36.0    & 21.3     & 6.6  & 53.2  & 31.0 & 2.9  & 2.89  & 0.43 & 79.8 & 82.8 \\
      % 0.48
      LLaVA-1.5     & 46.3    & 22.6        & 7.8    & 51.0   & 36.4    & 4.2 &  2.42  & - & 72.5 & 84.5  \\
      % CogVLM-Chat        & 73.6    & 87.4    & 70.0    & 85.0    & 80.0    & 40.0    & 2.51 & 0.49   & - & - & - & - & 55.00 \\
          LLaVA-RLHF    & 38.1 & 18.9    & 7.7 & 52.1 & 39.0  & 4.4 & 2.53 & 0.57 & 76.9 & 80.5  \\       
    RLHF-V  & 12.2 & 7.5 & 6.3  & 46.1 & 25.1  & 2.1 & 2.81  & 0.49 & 59.7 & - \\

      GPT-4V       & 13.6 & 7.3  & 4.6  & 67.1 & 30.7 &  2.6  & 3.49 & 0.28 &  - &  - \\
      Silkie & 25.3 & 13.9 & 5.4 & 55.8 & 29.0 & 2.0 & 3.01 & 0.41 & 84.9 & 82.1\\
        % Silkie-8K*\tablefootnote{Silkie-8K is trained on randomly sampled 8k data from silkie dataset} & 35.0 & 20.1 & 120.6 & 5.6 & 55.4 & 32.4 & 2.9 & 2.58 & 0.52 & 84.4 & 83.1\\
      \midrule
            % \textbf{Qwen w/ DPO(4K )} & 14.3 & 8.0 & 85.3 & 3.5 & 51.2 & 16.8 & 1.2 & 2.90 & 0.38 & 82.0 & -\\
    \multicolumn{11}{l}{\text{DPO}}  \\ \hline
    \text{w/ Qwen-VL} & 14.3 & 8.0  & 3.8 & 53.2 &  19.7  & 1.8  & 2.98 & 0.38 & 82.0 & 82.6\\ 
    \text{w/ LLaVA-1.5} & 6.7 & 3.6  & 2.8 & 47.8 & 15.5 & 1.6 & 2.58 & 0.50 & 79.3 & 84.5\\

    \hline
    \multicolumn{11}{l}{\text{HSA-DPO}}  \\ \hline
    % \text{w/ DPO} & 6.7 & 3.6  & 2.8 & 47.8 & 15.5 & 1.6 & 2.58 & 0.50 & 79.3 & 84.5\\
    % \text{w/ DPO} & 14.3 & 8.0  & 3.8 & 53.2 &  19.7  & 1.8  & 2.98 & 0.38 & 82.0 & 82.6\\
    \text{w/ Qwen-VL} & 11.0 & 5.5  & 3.7 & 52.4 &  19.0  & 1.6  & 3.07 & 0.34 & 82.4 & 82.9\\
    \text{w/ LLaVA-1.5} & \textbf{5.3} & \textbf{3.2}  & \textbf{2.1} & 47.3 &  \textbf{13.4}  & \textbf{1.2}  & 2.61  & 0.48 & 80.5 & 84.9\\
      % RLHF-V (800)   & 96.0 & 96.1 & 76.7 & 83.3  & 90.0 & 41.7  & 2.58 & 0.51  & 89.5 &  66.3 & 84.7 & 80.1 & \textit{79.85} \\
      % RLHF-V (1300 + QA)  & 92.8 & 95.3 & - & -  & - & -  & 2.47 & 0.53 (0.40) & 86.2 &  58.1 & 83.9 & 76.0 & \textit{79.85} \\
  % RLHF-V (1300 + QA w SFT)  & 86.2 & 92.4 & - & -  & - & -  & - & -  & 92.0 &  72.1 & 91.4 & 85.2 & - \\

   \bottomrule
    \end{tabular}
  }
\caption{Main experimental results on hallucination mitigation. Note that LLaVA Bench denotes LLaVA Bench in the wild\cite{llava}. POPE Adv. denotes POPE Adversarial\cite{pope}.}
  \label{tab:main result on mitigation}
\end{table*}

\subsection{Baselines}
For hallucination detection, we compare our method with GPT-4V\cite{gpt-4V}, Gemini\cite{gemini} and UNIHD\cite{unified} following the experiment settings of \cite{unified}.

For hallucination mitigation, we adopt a range of competitive hallucination mitigation methods as baselines: (1) InstructBLIP\cite{instructblip}; (2) LLaVA 1.5\cite{llava-1.5}; (3) Qwen-VL-Chat\cite{qwen-vl}; (4) GPT-4V\cite{gpt-4V}; (4) LRV~\cite{lrv}; (5) LLaVA-RLHF~\cite{llava-rlhf}; (6) RLHF-V\cite{rlhf-v}; (7) Silkie\cite{silkie}; (8) POVID~\cite{zhou2024povid}. More details about these baselines can be found in the related work section.

\subsection{Implementation Details}
For the detection model, we tune InternVL-Chat-Plus-v1.2 with LoRA~\cite{lora}. We take LLaVA-1.5-13B with zero-shot prompting as the rewriting model. For HSA-DPO, we fine-tune base models with LoRA.

\subsection{Main Results}
We report main results with respect to hallucination detection and mitigation, respectively.
\paragraph{Hallucination Detection.}
Table~\ref{tab:main results in detection} and \ref{tab:finegrainedresults} report the main results on hallucination detection benchmarks. We can draw the following conclusions. 
First, our detection model achieves the state-of-the-art results on MHaluBench on average, outperforming GPT-4V and Gimini.
Specifically, at the claim level, our detection model relatively surpasses UNIHD by 4.7\% in Mac. F1 score and GPT-4V Self-Check 2-shot by 8.1\% in Mac. F1 score. 
This improvement is consistently observed at the segment level as well. 
Second, on MFHaluBench, our detection model achieves an F1 score of 88.2\% in binary classification, and an accuracy of 74.3\% in Multi (fine-grained classification), outperforming GPT-4V 2-shot and LLaVA-1.6-34B 2-shot. 
\paragraph{Hallucination Mitigation.}
Table~\ref{tab:main result on mitigation} reports the main results on hallucination mitigation benchmarks. 
We can draw the following conclusions. 
First, HSA-DPO achieves state-of-the-art results on Object HalBench, outperforming the leading-edge closed-source LVLMs like GPT-4V.
Second, HSA-DPO reduces the hallucination of LLaVA-1.5, our base model, by 76.3\% for $\text{CHAIR}_{S} $ on Object HalBench and by 36.1\% for Hal on AMBER.
Third, compared to models trained on coarse-grained AI feedback (Silkie) or human-labelled fine-grained dataset (RLHF-V), our method gives superior results in hallucination mitigation. This demonstrates the effectiveness of using fine-grained AI feedback for detecting and mitigating hallucinations in LVLMs. 
Fourth, after preference learning, HSA-DPO is not affected by \emph{alignment tax}~\cite{askell2021general, ouyang2022training} and maintains its multi-modality capabilities, as evidenced by the results on the overall metric of MMHal-Bench and LLaVA Bench in the wild.
Lastly, to investigate the effect of our method on different base models, we train HSA-DPO on Qwen-VL-Chat\cite{qwen-vl}, where the hallucinated responses are generated by Qwen-VL-Chat. Both DPO and HSA-DPO results in Table~\ref{tab:main result on mitigation} indicate that our model gives a significant improvement over Qwen-VL-Chat, which demonstrate that our methodology can be applied to various LVLMs for mitigating hallucinations.
% \paragraph{Hallucination Severity Score.} 
% \Leilei{TODO.}
%We also compare our method with Silkie\cite{silkie}, which is also trained on Qwen-VL-Chat but with 80k coarse-grained preference dataset.
% \vspace{-5pt}
\subsection{Ablations}

    % \setlength{\tabcolsep}{1.3mm}
    % \begin{table}[t]
    %     \caption{Ablation results on detection model. C. denotes CHAIR, ObjHal denotes Object HalBench.}
    %     \label{tab:ablation of detection model}
    %     \small
    %     \centering
    %     % \resizebox{0.46\textwidth}{!}
    %     { % 0.8 表示宽度为文本宽度的 80%
    %     \begin{tabular}{cccc|ccc}
    %         \toprule
    %         \multirow{2}{*}{\textbf{Feedback}} & \multicolumn{3}{c}{\textbf{ObjHal}} & \multicolumn{3}{c}{\textbf{AMBER}} \\ \cline{2-7}
    %                                   & C.s$\downarrow$ & C.i$\downarrow$ 
    %                                   & C.$\downarrow$ & Cover$\uparrow$ 
    %                                   & {Hal$\downarrow$}   & {Cog$\downarrow$} \\ 
    %         \midrule
    %         Our pipeline & 5.3 & 3.2 & 2.1 & 47.3 & 13.4 & 1.2 \\
    %         w/o detection model & 42.1    & 20.3    & 7.6    & 52.1   & 32.4    & 4.0\\
    %         \bottomrule
    %     \end{tabular}
    %     }
    % \end{table}

    \setlength{\tabcolsep}{0.8mm}
    \begin{table}[t]

        \small
        \centering
        % \resizebox{0.46\textwidth}{!}
        { % 0.8 表示宽度为文本宽度的 80%
        \begin{tabular}{ccc|ccccc}
            \toprule
            \multirow{2}{*}{\textbf{Methods}} & \multicolumn{2}{c}{\textbf{ObjHal}} & \multicolumn{4}{c}{\textbf{AMBER}} &
            \multirow{2}{*}{\textbf{HS$\downarrow$}} \\ 
            \cline{2-7}
                                      & C.s$\downarrow$ & C.i$\downarrow$ 
                                      & C.$\downarrow$ & Cover.$\uparrow$ 
                                      & {Hal.$\downarrow$}   & {Cog.$\downarrow$} \\ 
            \midrule
            Ours & 5.3 & 3.2 & 2.1 & 47.3 & 13.4 & 1.2 & 0.60\\
            w/o Detection & 42.1    & 20.3    & 7.6    & 52.1   & 32.4    & 4.0 & 0.79\\
            w/o HSA & 6.7    & 3.6    & 2.8    & 47.8  & 15.5    & 1.6 & 0.65\\
            w/o FineGrained & 17.0    & 8.9   & 5.0    & 53.6   & 27.3    & 1.7 & 0.67\\
            \bottomrule
        \end{tabular}
        }
        \caption{Ablation results on Object HalBench (ObjHal) and AMBER. HS denotes Hallucination Severity rated by GPT-4V. C.s, C.i, C. are short for CHAIR\(_{S}\), CHAIR\(_{I}\), CHAIR.}
        \label{tab:ablation of detection model}
    \end{table}

We conduct ablation studies to validate the effectiveness of each design of our method. The results are reported in Table~\ref{tab:ablation of detection model}, where "w/o Detection" represents instead of detecting hallucinations by the detection model, we directly rewrite the hallucinated response into non-hallucinated one to construct preference dataset. "w/o HSA" represents that we do not incorporate hallucination severity into preference optimization and use the vanilla DPO. "w/o FineGrained" means we use coarse-grained GPT-4V feedback.

From the table, we have the following conclusion. First, 
without detection model identifying hallucinations, model performance seriously degrades on both benchmarks, highlighting the necessity of our detect-then-rewrite pipeline. Second, HSA-DPO demonstrate robust improvement over vanilla DPO in mitigating hallucinations across all hallucination metrics, relatively reducing by 20.8\% for $\text{CHAIR}_{S} $ on Object HalBench, by 13.5\% for Hal on AMBER and by 7.6\% on Hallucination Severity. Third, fine-grained feedback outperforms coarse-grained one from GPT-4V, relatively reducing by 68.8\% for $\text{CHAIR}_{S} $ on Object HalBench, by 50.9\% for Hal on AMBER, which reveals the efficacy of fine-grained granularity.
% we alter our fine-grained feedback with feedback from GPT-4V filtered from Silkie~\cite{silkie}. Results in Table ~\ref{tab:ablation of detection model} indicates that our feedback is comparable to feedback from GPT-4V.
\subsection{Analyses}
\label{sec:analy}
\paragraph{Annotation Efficiency and Cost.}
Our method constructs preference dataset using AI feedback through a detect-then-rewrite pipeline. Table \ref{tab:complexity} compares the annotation efficiency and cost with those of GPT-4V and human annotation. While our method needs 5.7 hours to train models, it only requires 25\% and 62.5\% of the time needed by human annotation and GPT-4V, respectively. Regarding annotation cost, our method incurs only \$600, which is 12.5\% and 37.5\% of the cost needed by human annotation and GPT-4V, respectively. These results underscore the superior annotation efficiency and cost-effectiveness of our approach compared to other methods.

\begin{figure}[t]
  % \centering
  \includegraphics[width=0.47\textwidth]{image/new_merged_image_with_labels.jpg}
  \caption{Effect of scaling preference dataset (Figure A) and different hallucination types (Figure B).}
  \label{fig:scaleup}
\end{figure}

\begin{table}[t]
    \centering
    \begin{tabular}{cccc}
    \toprule
      \textbf{Feedback}   & \textbf{Train Time} & \textbf{Efficiency} & \textbf{Cost}\\
    \midrule
      Human   &  0 & 20 s+  & \$4800 \\  
      GPT-4V & 0  & 8 s & \$1600 \\
      Our pipeline & 5.7 h & 5 s & \$600 \\
    \bottomrule
    \end{tabular}
     \caption{Annotation Efficiency and Cost. Efficiency is annotation time per sample and Cost is for collecting 16k preference datapoints. }
    \label{tab:complexity}
\end{table}

\begin{figure}[t]
  \centering
  \includegraphics[width=0.47\textwidth]{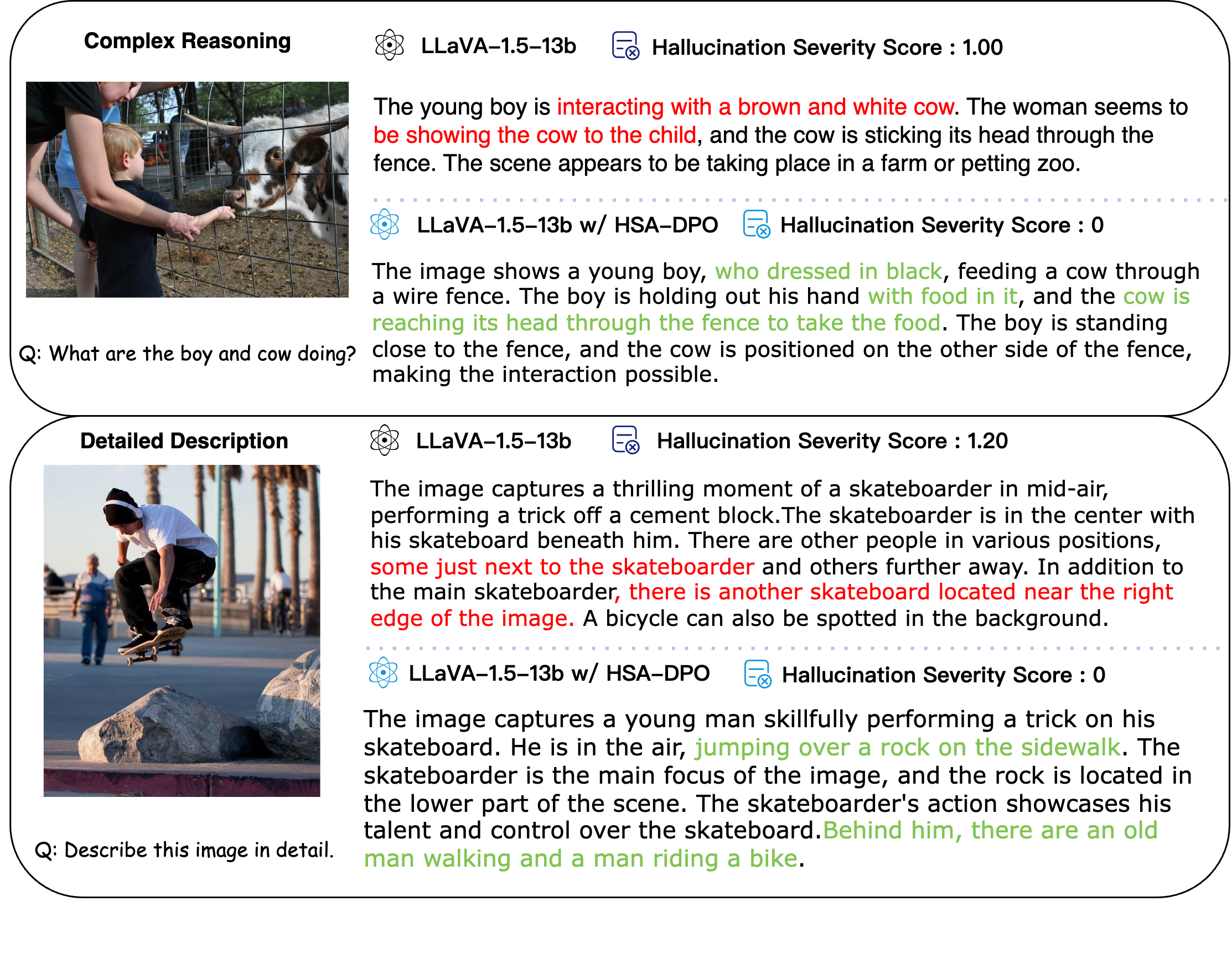}
  \caption{Qualitative results of different models on VCR and DDG. Correct answers, factual hallucinations are highlighted in red and green respectively.}
  \label{fig:goodcases}
\end{figure}
% \vspace{-5pt}

\paragraph{Effect of Scaling Preference Dataset.}
We conduct experiments to investigate the effect of scaling the preference dataset on Object HalBench. As depicted in Figure \ref{fig:scaleup}, we observe that with the increasing size of dataset, the hallucination rate of LLaVA with HSA-DPO shows a rapid and consistent decrease. When the size of the training dataset reaches 8k, our model surpasses GPT-4V on both $\text{CHAIR}_{S} $ and $\text{CHAIR}_{I} $, highlighting the value of our pipeline in enabling cost-effective annotation of preference datasets.

\paragraph{Effect on Different Hallucination Types.}
To evaluate the effect of our method on different hallucination types, we conduct experiments on Amber benchmark and report the F1 scores on all types, i.e., object existence, attribute, state, number, action and relation. Figure \ref{fig:scaleup} shows HSA-DPO with LLaVA-1.5 outperforms LLaVA-1.5 across all hallucination types. We also find that our method are more effective in improving number and existence types (object hallucinations) . However, we observe limited improvement in the relation and action types (relationship hallucinations). This is likely due to the scarcity of relationship hallucinations in preference dataset, as well as the inherent complexity of addressing relationship hallucinations compared to object ones.
% The enhancements are particularly notable in the dimensions of object existence and attributes.
%Here, 'states' furthers the category of 'attributes' to include such characteristics as color and shape, 'numbers' refer to the numbers of the object within the image content, and 'actions' refer to the activities of humans or animals within the image content. More details about these hallucination types can refer to the original paper.

\subsection{Case Studies}

In Figure \ref{fig:goodcases}, we qualitatively compare model performance on VCR and DDG. In VCR case, LLaVA struggles  to recognize the key action of the boy feeding the cow. After HSA-DPO training, model accurately answers the question with more details. For DDG, LLaVA makes serious errors about nearby people and skateboard. However, model with HSA-DPO provides a precise description, accurately identifying relationships and objects in the image.
% \vspace{-5pt}
\section{Conclusion}
In this work, we propose detecting and mitigating hallucinations in LVLMs via fine-grained AI feedback. We begin with generating sentence-level hallucination annotation dataset via AI feedback. Then, a detect-then-rewrite pipeline is used to more cost-effectively construct preference dataset at scale. Lastly, HSA-DPO is introduced to incorporates hallucination severity into preference learning for prioritizing the mitigation of critical hallucinations. Extensive experiments have demonstrated the effectiveness of our method.

\section{Acknowledgments}
This work was supported by the National Natural Science Foundation of China (62441605, 62376243, 62037001, U20A20387), and the Starry Night Science Fund of Zhejiang University Shanghai Institute for Advanced Study (SN-ZJU-SIAS-0010). This work was also supported in part by the "Pioneer" and "Leading Goose" R\&D Program of Zhejiang (No.2024C01142). This work was supported by Alibaba Group
through Alibaba Innovative Research Program.

\bibliography{aaai25}

\end{document}